\pdfoutput=1
\documentclass[11pt]{article}
\usepackage{graphicx}
\usepackage[preprint]{acl}
\usepackage{times}
\usepackage{latexsym}
\usepackage[T1]{fontenc}
\usepackage[utf8]{inputenc}
\usepackage{microtype}
\usepackage{inconsolata}
\usepackage{multirow}
\usepackage{booktabs}

\usepackage{amsmath,amsfonts,bm}

\def\eqref#1{equation~\ref{#1}}

\def\1{\bm{1}}

\def\va{{\bm{a}}}

\def\vf{{\bm{f}}}

\def\vi{{\bm{i}}}

\def\vo{{\bm{o}}}
\def\vp{{\bm{p}}}
\def\vq{{\bm{q}}}

\def\mA{{\bm{A}}}

\def\mI{{\bm{I}}}

\def\mP{{\bm{P}}}
\def\mQ{{\bm{Q}}}

\def\mS{{\bm{S}}}
\def\mT{{\bm{T}}}

\DeclareMathAlphabet{\mathsfit}{\encodingdefault}{\sfdefault}{m}{sl}
\SetMathAlphabet{\mathsfit}{bold}{\encodingdefault}{\sfdefault}{bx}{n}

\def\gA{{\mathcal{A}}}

\def\gD{{\mathcal{D}}}

\def\gI{{\mathcal{I}}}

\def\gL{{\mathcal{L}}}

\def\gQ{{\mathcal{Q}}}

\def\gS{{\mathcal{S}}}

\def\sD{{\mathbb{D}}}

\def\sT{{\mathbb{T}}}

\DeclareMathOperator*{\llm}{LLM}
\DeclareMathOperator*{\retrieval}{Retrieval}
\DeclareMathOperator*{\rank}{Rerank}
\DeclareMathOperator*{\vencoder}{Vision-Enc}
\DeclareMathOperator*{\clipencoder}{VL-Enc}
\DeclareMathOperator*{\simi}{Sim}
\DeclareMathOperator*{\lvlm}{LVLM}
\DeclareMathOperator*{\prompt}{Prompt}
\DeclareMathOperator*{\concat}{Concat}
\DeclareMathOperator*{\icl}{icl}

\title{Visual In-Context Learning for Large Vision-Language Models}

\author{Yucheng Zhou$^{1}$, Xiang Li$^{2}$, Qianning Wang$^{3}$, Jianbing Shen$^{1}$\thanks{~Corresponding author.}\\
         $^{1}$ SKL-IOTSC, CIS, University of Macau \\
         $^{2}$ Tianjin University, $^{3}$ Nanjing Audit University \\
         {\tt yucheng.zhou@connect.um.edu.mo, jianbingshen@um.edu.mo}
         }

\begin{document}
\maketitle
\begin{abstract}
In Large Visual Language Models (LVLMs), the efficacy of In-Context Learning (ICL) remains limited by challenges in cross-modal interactions and representation disparities. To overcome these challenges, we introduce a novel Visual In-Context Learning (VICL) method comprising Visual Demonstration Retrieval, Intent-Oriented Image Summarization, and Intent-Oriented Demonstration Composition. Our approach retrieves images via ``Retrieval \& Rerank'' paradigm, summarises images with task intent and task-specific visual parsing, and composes language-based demonstrations that reduce token count and alleviate cross-modal interaction problem. Experimental evaluations on five visual reasoning datasets demonstrate the effectiveness of our method. Moreover, our extensive experiments leverage information flow analysis to elucidate the effectiveness of our method, and investigate the impact of length and position of demonstrations for LVLM. The use of in-context unlearning further shows promise in resetting specific model knowledge without retraining. 
\end{abstract}

\section{Introduction}
Large Language Models (LLMs) exhibit impressive reasoning abilities across various natural language tasks \citep{gpt3,Touvron2023LLaMAOA}. Researchers are actively investigating the extension of LLM capabilities to address challenges in the visual domain by integrating LLMs with vision models \citep{zhu2023minigpt,Bai2023QwenVLAF}. This endeavor has given rise to the development of Large Visual Language Models (LVLMs). LVLMs are designed to seamlessly fuse information from both images and text, enabling them to tackle more intricate tasks that demand a profound comprehension of both modalities \citep{2023GPT4VisionSC,zhu2023minigpt}.

\begin{figure}[t]
    \centering
    \includegraphics[width=\linewidth]{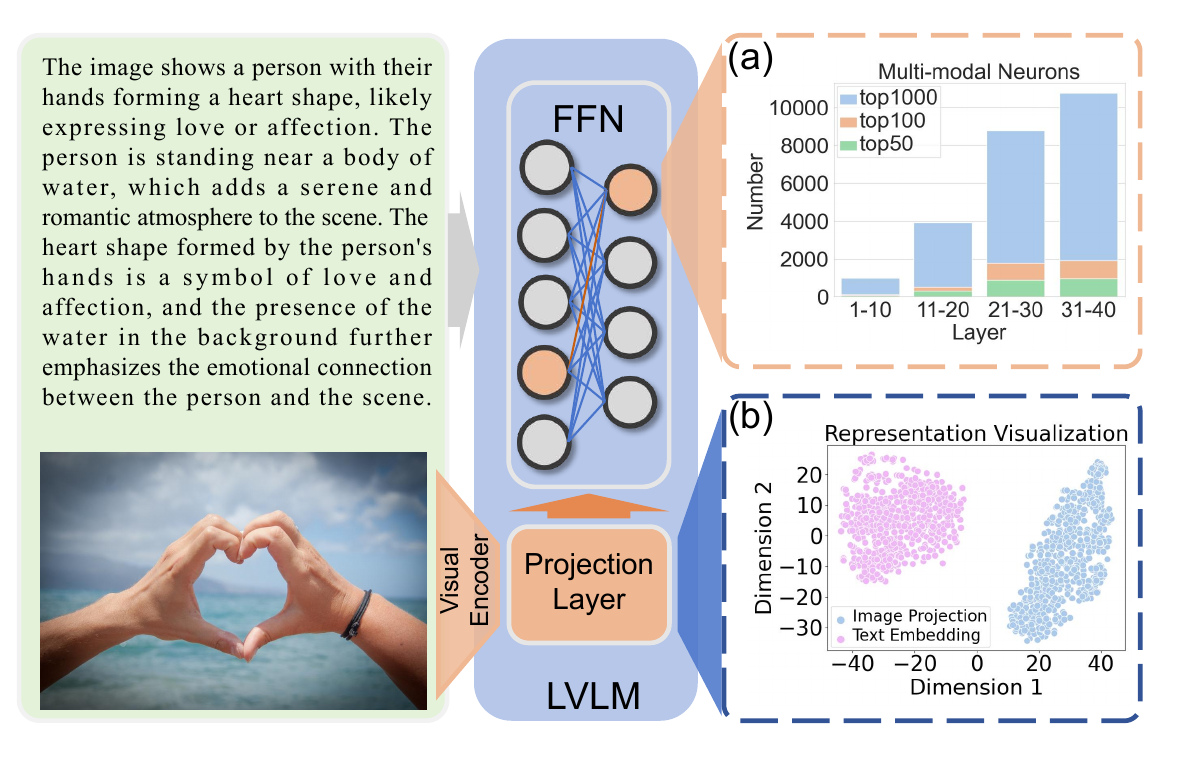}
    \vspace{-0.8cm}
    \caption{\small Illustrating Cross-Modal Challenges in LVLMs: (a) Distribution of multi-modal interaction neurons. (b) The distinct spaces are occupied by visual features and text embeddings before passing into LVLMs.}
    \label{fig:intro} 
    \vspace{-0.5cm}
\end{figure}

The LLM has a remarkable capability, known as In-Context Learning (ICL), which involves providing LLMs with a limited amount of labeled data as demonstrations to improve their reasoning ability \cite{gpt3, Dong2023ASF, cot2}. This approach can significantly enhance the performance of LLMs in various NLP tasks, such as translation \cite{Garca2023TheUE, Moslem2023AdaptiveMT}, sentiment classification \cite{sentiment}, and question-answering \cite{sentiment, ICLQA1}. In the ICL, LLMs can flexibly adjust their behavior based on the provided context, allowing them to understand and perform tasks with few labeled data. The success of ICL has motivated research into extending the ICL capabilities to LVLMs. However, some studies \cite{deeplayer, deeplayer2} find that while LVLMs have ICL capabilities, they are not as pronounced as those observed in LLMs. Two factors lead to this difference: (1) As shown in Figure~\ref{fig:intro}(a), as observed in previous research \cite{Neural1, Neural2}, visual-language interactions occur at deeper layers in LVLMs, which highlights the difficulty of cross-modal interactions. In ICL, label words aggregate information in shallow layers and subsequently distribute it in deeper layers \cite{wordsanchor}. Consequently, the issue of cross-modal interactions significantly impacts the ICL capabilities of LVLMs. (2) From Figure~\ref{fig:intro}(b), our analysis reveals that visual features and LLM embeddings occupy distinct spaces in LVLMs. This observation underscores the inherent cross-modal gap present in LVLMs.
While many works \cite{Min2022RethinkingTR, Lu2023AreEA} are dedicated to enhancing the ICL capabilities of LLMs, the challenges faced by LVLMs in this regard differ substantially. This discrepancy arises from the difficulty in cross-modal interactions and inherent disparities in representation spaces within LVLMs, which impose limitations on their ICL performance. 

In this study, we present a novel Visual In-Context Learning (VICL) method to enhance the ICL capability of LVLMs. VICL comprises Visual Demonstration Retrieval, Intent-Oriented Image Summarization, and Intent-Oriented Demonstration Composition.
For Visual Demonstration Retrieval, we employ a pre-trained image encoder as a retriever to search for relevant candidate images for the provided image. Subsequently, we rerank the retrieved candidates using textual descriptions of the provided image.
Moreover, LVLMs perform Intent-Oriented Image Summarization, automatically extracting image summary with task intent and task-specific visual parsing from image-label pairs. 
In addition, Intent-Oriented Demonstration Composition uses language cues to create demonstrations that enhance ICL of LVLMs, replacing images with image summary in demonstrations. 
Our method not only boosts in-context learning but also introduces in-context unlearning, allowing models to discard or reset specific knowledge through demonstration.
The substitution of images with visual summaries significantly reduces the token count, enabling the concatenation of more demonstrations without encountering token limitations. In comparison to conventional visual-language interactions in standard visual ICL approaches, our method solely relies on language interactions to facilitate effective demonstration understanding.

Our experiments across five image reasoning datasets evaluate our method's effectiveness, comparing LVLM performance using our approach against a baseline method. 
Moreover, we employed information flow for interpretative analysis, to verify the effectiveness of our method. Furthermore, examined the influence of demonstrations and their sequence length on LVLM's ICL capability. We investigate the importance of intent-oriented image summaries and the impact of demonstrations order. In addition, we explore the application of in-context unlearning, demonstrating its feasibility for unlearning scenarios without additional training.

\section{Related Work}
\subsection{Large Vision-Language Models}

Large Vision-Language Models (LVLMs) are designed to comprehend and generate content across vision and language modalities, allowing them to perform tasks that involve understanding and generating not only text, but also information in visual forms \cite{LVLMsurvey}.

LVLM can be broadly categorized into two main types, according to the output modalities: visual understanding and visual generation. Visual understanding models are capable of comprehending visual modality information provided to them and generating textual responses, enabling them to accomplish tasks such as image captioning, image question answering \cite{zhu2023minigpt, 2023GPT4VisionSC, Alayrac2022FlamingoAV}, video understanding \cite{VLT5}, video captioning \cite{VideoCon}, etc. The typical structure of these models involves integrating the visual encoders based on transformer architecture (like Clip\cite{RadfordKHRGASAM21}) into a large language model.

On the other hand, visual generation models are equipped with visual decoders, enabling the decoding of feature vectors into images or videos. They have shown the ability to create high-quality outputs in generative tasks, such as generating text, images, and videos \cite{DALLE2, Imagen, InternLM-XComposer}.

However, since the LVLM has strong capabilities, it memorizes much unnecessary knowledge. In certain scenarios involving security or privacy concerns, it becomes imperative to selectively erase specific knowledge acquired by machine learning models \cite{Goldsteen2020DataMF}. Unlike conventional databases where information is explicitly stored in tabular forms, the entirety of a model's acquired knowledge is implicitly embedded within its parameters. Consequently, the challenge arises of accurately expunging unwanted information without necessitating a complete retraining of the model, thereby minimizing interference with other retained knowledge. This intricate problem is addressed by a collective set of techniques known as machine unlearning \cite{Bourtoule2019MachineU, munl1, munl2, munl3}. 

\subsection{In-Context Learning}
In-Context Learning exemplifies a paradigm where model weights require no optimization; rather, adjusting the model input (adding context) leads to correct output generation \cite{Dong2023ASF}. An in-context learning prompt typically consists of two components: demonstration and new query. Demonstrations comprise multiple question-answer pairs, each presenting a complete question and its corresponding answer, while new queries involve inquiries posed to the model. Due to the emergent ability in large language models \cite{Lu2023AreEA}, they can to some extent reference demonstrations to answer new questions \cite{Min2022RethinkingTR}. With the advantage of not necessitating fine-tuning of model parameters, in-context learning has become a popular paradigm for applying large language models. 

The inherent black-box nature of deep neural models renders the reasons behind the efficacy of in-context learning even more challenging to elucidate \cite{iclexp1, iclexp2, iclexp3, iclexp4, iclexplanation}. One of the most widely accepted theoretical explanations at present is that when the pre-training text has long-range coherence, if the demonstrations in the prompt share potential concepts, in-context learning ability will emerge \cite{iclexplanation}. 

Motivated by in-context learning, the in-context unlearning \cite{Pawelczyk2023InContextUL} emerges as a promising solution, which specifically applies the in-context learning paradigm without updating any model parameters, making it suitable for large language models. The framework leverages a combination of incorrectly and correctly labeled examples from training datasets. By analyzing and understanding the nuances within these discrepancies, a unique prompt is constructed for each instance. This tailored prompt aims to highlight specific challenges posed by the labeling discrepancies, encouraging the model to refine its predictions during inference.

\section{Visual In-Context Learning}
\begin{figure*}[t]
    \centering
    \includegraphics[width=\linewidth]{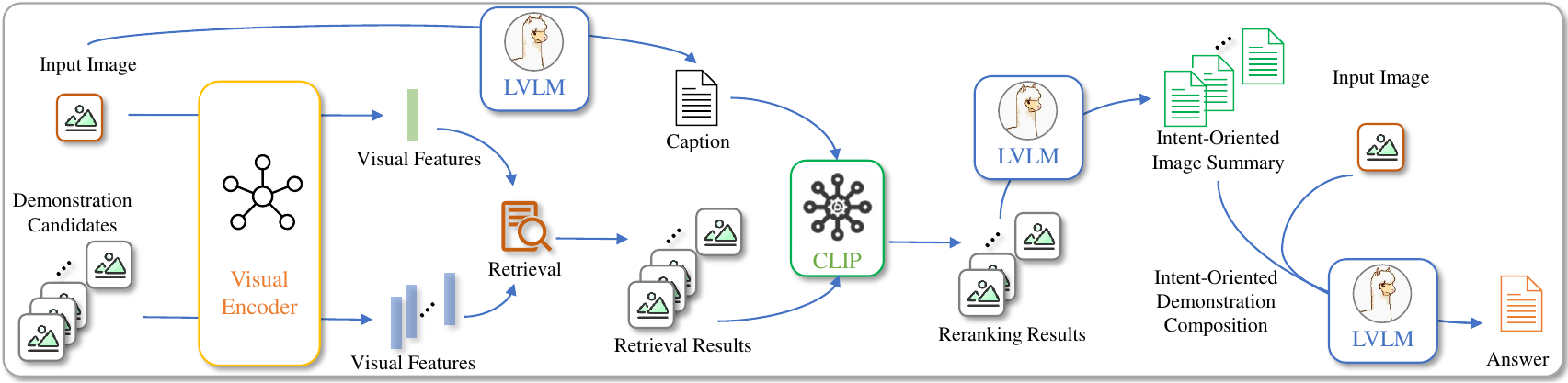}
    \caption{\small Overview of our Visual In-Context Learning (VICL) method. The Visual Encoder is used to encode images for retrieval, and CLIP is used for cross-modal reranking of images and caption; LVLM is used to generate caption for input images, generate intent-oriented image summaries, and predict answer based on the composed prompt.}
    \label{fig:method}
\end{figure*}

In this section, we elaborate on our approach VICL, which comprises three core components: Visual Demonstration Retrieval, Intent-Oriented Image Summarization, and Intent-Oriented Demonstration Composition. The pipeline of VICL is shown in Figure~\ref{fig:method}.

\subsection{Background}
Generative language models could self-supervised learn knowledge from pre-training corpora \cite{gpt1,gpt2,t5}. As the scale of model parameters and pre-training corpora expands, researchers have observed the emergent ability in large language models, enabling them to provide accurate answers merely by adjusting the input prompt without fine-tuning model parameters \cite{Wei2022EmergentAO,Wei2022ChainOT,Fu2022ComplexityBasedPF}. 

The large language model is abstracted as a function denoted as $\llm( \cdot )$. Given an input prompt, denoted as $\vp$, it generates the corresponding output, denoted as  $\vo$. In the most common case of question-answering, the prompt $\vp$ is exactly the question $\vq$ raised by the user. And the output $\vo$ is expected to be the answer $\va$ to the question $\vq$.
\begin{align}
\label{eq:iclParadigm}
\llm(\vo|\vp) &= \llm(\va|\vq)
\end{align}

However, under the in-context learning paradigm, the input prompt $\vp_{\icl}$ is carefully designed:
\begin{align}
\notag \llm(\va|\vp_{\icl}) &= \llm(\va|\sD \oplus \vq) \\
\sD &= \sT(\oplus_{j}^{n}(\hat{\vq_j} \oplus \hat{\va_j})),
\end{align}
where the input prompt (denoted as $\vp_{\icl}$) is formed by concatenating (denoted as $\oplus$) demonstrations $\sD$ and query $\vq$, the current question raised by the user. $\sD$ is composed of number $n$ sets of complete questions($\hat{\vq_j}$) and answers($\hat{\va_j}$), spliced together through a fixed template $\sT(\cdot)$.

Considering the established efficacy of in-context learning on large language models, it was intuitive to extend the in-context learning approach to large visual-language models upon its emergence \cite{vicl1, vicl2, vicl3, vicl4}. In the context of a visual question answering task, the formulation of in-context learning on large visual-language model can be delineated as follows:
\begin{align}
\notag \lvlm(\va|\vp_{\icl}) &= \lvlm(\va|\sD \oplus \vi \oplus \vq) \\
\sD &= \sT(\oplus_{j}^{n}(\hat{\vq_j} \oplus \hat{\vi_j} \oplus \hat{\va_j})),
\end{align}
where the visual input prompt, denoted as $\vp_{\icl}$, is created by concatenating demonstrations $\sD$, query image $\vi$, and query text $\vq$. $\sD$ consists of $n$ sets of complete questions ($\hat{\vq_j}$), images ($\hat{\vi_j}$), and answers ($\hat{\va_j}$), which are combined by a fixed template $\sT(\cdot)$.

\subsection{Visual Demonstration Retrieval}
The Visual Demonstration Retrieval (VDR) component is the first step in our VICL method, designed to identify suitable samples for ICL as demonstrations. The goal of VDR is to discern and choose visual demonstrations that bear the utmost relevance to the current task. This process capitalizes on both the visual features of images and their accompanying textual descriptions. Following de facto ``retrieval \& rerank'' paradigm \cite{Zhou0GTXLJJ23}, VDR comprises these two phases.

\paragraph{Visual Retrieval.}
Given an image $\mI$, our goal is to find a set of candidate demonstration images $\gD$ = $\{\mI_1, \mI_2, \cdots, \mI_n\}$ that are relevant to $\mI$. This is achieved by employing a pre-trained image encoder, $\vencoder$, that maps images into a high-dimensional feature space, i.e.,
\begin{align}
\label{eq:vencoder}
\vf &= \vencoder(\mI),\\
\vf_i &= \vencoder(\mI_i), i \in \{1, 2, \cdots, n\},
\end{align}
where $\vf$ is the feature vector representing the embedding of the image $\mI$, and $\vf_i$ represents the embedding of the image $\mI_i$ in the dataset. $\vencoder$ denotes ViT \cite{DosovitskiyB0WZ21}.
The retrieval operation is defined as:
\begin{align}
\gD_q = \retrieval(\mI, \gD \mid \vencoder),
\end{align}
The $\retrieval$ function aims to select the top-$n$ images from $\gD$ whose embeddings are most similar to the embedding of $\mI$, based on a similarity metric $\simi$. This approach ensures that the selected images $\gD_q$ are those that are most relevant to the query image in terms of visual features encoded within the high-dimensional feature space. The similarity metric is defined as:
\begin{align}
\label{eq:similarity}
\simi(\vf, \vf_i) = \frac{E(\vf) \cdot E(\vf_i)}{\|E(\vf)\| \|E(\vf_i)\|},
\end{align}
where $(\cdot)$ is the dot product between two vectors.

\paragraph{Cross-Modal Reranking.}
After the visual retrieval, we obtain a set of candidate images $\gD_q$. However, to ensure that the selected demonstrations are not only visually similar but also semantically relevant to $\mI$, we employ a reranking step using textual descriptions. We use a large vision language model, $\lvlm$, to generate an image description $\mT_q$ for the image $\mI$. The reranking process adjusts the initial rankings based on the semantic similarity between $\mT_q$ and $\mI_i$, $\mI_i \in \gD_q$, which is computed by a pre-trained image-text model $\clipencoder$:
\begin{align}
\label{eq:rerank}
\hat{\gD_q} = \rank(\gD_q, \mT_q \mid \clipencoder),
\end{align}
where $\hat{\gD_q}$ denotes the reranked set of demonstration images. $\clipencoder$ refers to CLIP \cite{RadfordKHRGASAM21}.
This dual-stage approach allows us to harness the complementary strengths of visual and textual modalities, ensuring that the chosen visual demonstrations are not only visually pertinent but also contextually aligned with the query image's task-specific requirements.

\subsection{Intent-Oriented Image Summarization}
Intent-Oriented Image Summarization (IOIS) aims to simplify LVLM's ICL problem by generating a visual content summary from a task intent perspective. This summarization process focuses on exploring the relationship of a given reference image, question and answer triplet, and generates an image summary encapsulating the task intent and the task-specific visual parsing.

\paragraph{ICL in LLM.} For LLMs, given the reference question-answering pair $\{\hat{\gQ}, \hat{\gA}\}$ and the input question $\mQ$, the ICL problem can be formalized as:
\begin{align}
    P(\mA \mid \mQ)\!=\!P(\mA \mid \mQ, \mT)\!\times\!P(\mT \mid \{\hat{\gQ}, \hat{\gA}\}),
\end{align}
where $\mA$ is the predicted answer; $\mT$ is task intent; the model needs to infer the task intent from given reference question-answer pairs to accurately respond to a new question.

\paragraph{ICL in LVLM.} In LVLMs, the ICL problem extends to:
\begin{align}
    \notag P(\mA \mid \mQ, \mI) &= P(\mA \mid \mI, \mQ, \mT, \mP) \\ \notag &\times P(\mT \mid  \{\hat{\gI}, \hat{\gQ}, \hat{\gA}\}) \\ &\times P(\mP \mid  \{\hat{\gI}, \hat{\gQ}, \hat{\gA}\}),
\end{align}
where the LVLM must first deduce the task intent $\mT$ and image parsing strategy $\mP$ from the reference image, question, and answer triplet $\{\hat{\gI}, \hat{\gQ}, \hat{\gA}\}$ before analyzing the content of the given image based on these insights.

\paragraph{VICL.} 
Our IOIS method significantly simplifies this process by pre-generating a visual content summary that embodies both the task intent and the image parsing approach. This summary is produced by concatenating a carefully designed prompt with the given reference image, question, and answer, and then inputting this into the LVLM. The output is a summary that not only describes the image but and encapsulates task intent and task-specific visual parsing. Our approach can be represented as:
\begin{align}
    P(\mA\!\!\mid\!\!\mQ,\!\mI)\!=\!P(\mA\!\!\mid\!\!\mI,\!\mQ,\!\gS)\!\!\times\!\!P(\gS\!\!\mid\!\!\{\hat{\gI},\!\hat{\gQ},\!\hat{\gA}\})
\end{align}
where $\gS$ denotes the set of intent-oriented visual summarization for all reference images. This formulation demonstrates how our method modifies the LVLM's ICL challenge by replacing the direct analysis of images with the interpretation of summarizations that are pre-aligned with the task's intent and preferred image parsing methodology.

\paragraph{Implementation of IOIS.}
The specific procedure for generating the Intent-Oriented Image Summarization involves constructing a prompt that integrates the demonstration image with its corresponding label, underpinned by the task's intent and image parsing preferences. This prompt is then input into the LVLM to produce the summarization. The process can be mathematically function as:
\begin{align}
    \notag \mS_i = \lvlm&(\prompt(\hat{\mI_i}, \hat{\mQ_i}, \hat{\mA_i}), \\
    \notag \text{where~} &\gS = \{\mS_1, \mS_2, \cdots, \mS_l\}, \\
    &\hat{\mI_i}, \hat{\mQ_i}, \hat{\mA_i} \in \{\hat{\gI}, \hat{\gQ}, \hat{\gA}\},
\end{align}
where $l$ is the number of reference examples.
$\prompt$ is a function that formulates the input for the LVLM, encapsulating the demonstration image and label along with explicit cues about the task's intent and the approach to image parsing.

This approach ensures that the LVLM's ICL process is primed with a context that significantly lowers the cognitive load associated with cross-modal mapping, allowing the model to focus on reasoning within a linguistic framework that is inherently more aligned with its training. This strategic simplification not only enhances the efficiency of the LVLM's ICL capabilities but also reduces the complexity associated with direct image analysis.

\subsection{Intent-Oriented Demonstration Composition}
Intent-Oriented Demonstration Composition (IODC) aims to effectively integrate the generated image summaries $\mS_i$ with corresponding questions $\mQ_i$ and answers $\mA_i$ into a coherent demonstration for the LVLM.

\paragraph{Composition of Demonstrations.} The IODC process involves the assembly of each image summary $\mS_i$ with its corresponding question $\mQ_i$ and answer $\mA_i$ into a single, unified demonstration. This is achieved through the concatenation of these elements in a manner that preserves the logical and semantic coherence necessary for effective ICL. Formally, the process can be represented as:
\begin{align}
    \bar{\gD_i} = \concat(\mS_i, \mQ_i, \mA_i),
\end{align}
where $\concat$ is the concatenation operation, and $\bar{\gD_i}$ represents the $i$-th demonstration composed of the image summary, question, and answer triplet. This operation is performed for each set of $\mS_i$, $\mQ_i$, and $\mA_i$, resulting in a collection of demonstrations:
\begin{align}
    \bar{\gD} = \{\bar{\gD_1}, \bar{\gD_2}, \cdots, \bar{\gD_n}\},
\end{align}
where $\bar{\gD}$ denotes the complete set of demonstrations ready for presentation to the LVLM.

\paragraph{Enhancing ICL with Demonstration Composition.} 
By replacing original images $\mI_i$ with intent-oriented visual summaries $\mS_i$, we significantly reduce the complexity and token count inherent in direct image processing. This reduction allows for the inclusion of a larger number of demonstrations within the LVLM's token limit, thereby enriching the context available for ICL without overwhelming the model with excessive information, i.e.,
\begin{align}
    \mA = \lvlm(\bar{\gD}, \mI, \mQ).
\end{align}
The IODC methodology facilitates a shift in LVLM processing from a reliance on direct visual inputs to an emphasis on linguistic representations of visual content, grounded in the task's intent. This shift not only streamlines the ICL process by minimizing the token count but also aligns the demonstrations more closely with the LVLM's linguistic processing capabilities. 

\subsection{Information Flow Analysis}
Following \citet{wordsanchor}, we analyze the information flow of VICL in the LVLM. To calculate saliency score of each element in attention matrix, we employ Taylor expansion \cite{MichelLN19}:
\begin{align}
I_l=\sum_h\left|\mA_{h, l}^{\top} \frac{\partial \gL(x)}{\partial \mA_{h, l}}\right|
\end{align}
where $h$ and $l$ represent different attention heads and transformer layers, respectively. $\gL(x)$ is the loss function for the task.
Furthermore, we define four different information flow significance scores as follows:
\begin{align}
\notag S_{wp} &= \frac{\sum_{(i,j) \in C_{wp}} I(i,j)}{|C_{wp}|}, \\
C_{wp} &= \{(p_k, j) : k \in [1, C], j < p_k\}\\
\notag S_{pq} &= \frac{\sum_{(i,j) \in C_{pq}} I_l(i,j)}{|C_{pq}|},\\
C_{pq} &= \{(q, p_k) : k \in [1, C]\}\\
S_{vq} &= \frac{\sum_{(i,j)\in C_{vq}} I_l(i,j)}{|C_{vq}|}, C_{vq} = \{(q, v)\}\\
\notag S_{ww} &= \frac{\sum_{(i,j)\in C_{ww}} I_l(i,j)}{|C_{ww}|}, \\
C_{ww} &= \{(i,j) : j < i\}\!-\!C_{wp}\!-\!C_{pq}\!-\!C_{vq}
\end{align}
where $p_k$, $C$, $q$ and $v$ represent the label words, the total number of label words, the target position and the input image, respectively. $S_{wp}$ denotes the significance of information flow from the image summaries to label words; $S_{pq}$ represents the significance of information flow from label words to the target position; $S_{vq}$ signifies the significance of information flow from label words to the input image part; $S_{ww}$ indicates the significance of the information flow amongst all words, excluding influences represented by $S_{wp}$, $S_{pq}$, and $S_{vq}$.

\begin{figure}[t]
    \centering
    \includegraphics[width=1\linewidth]{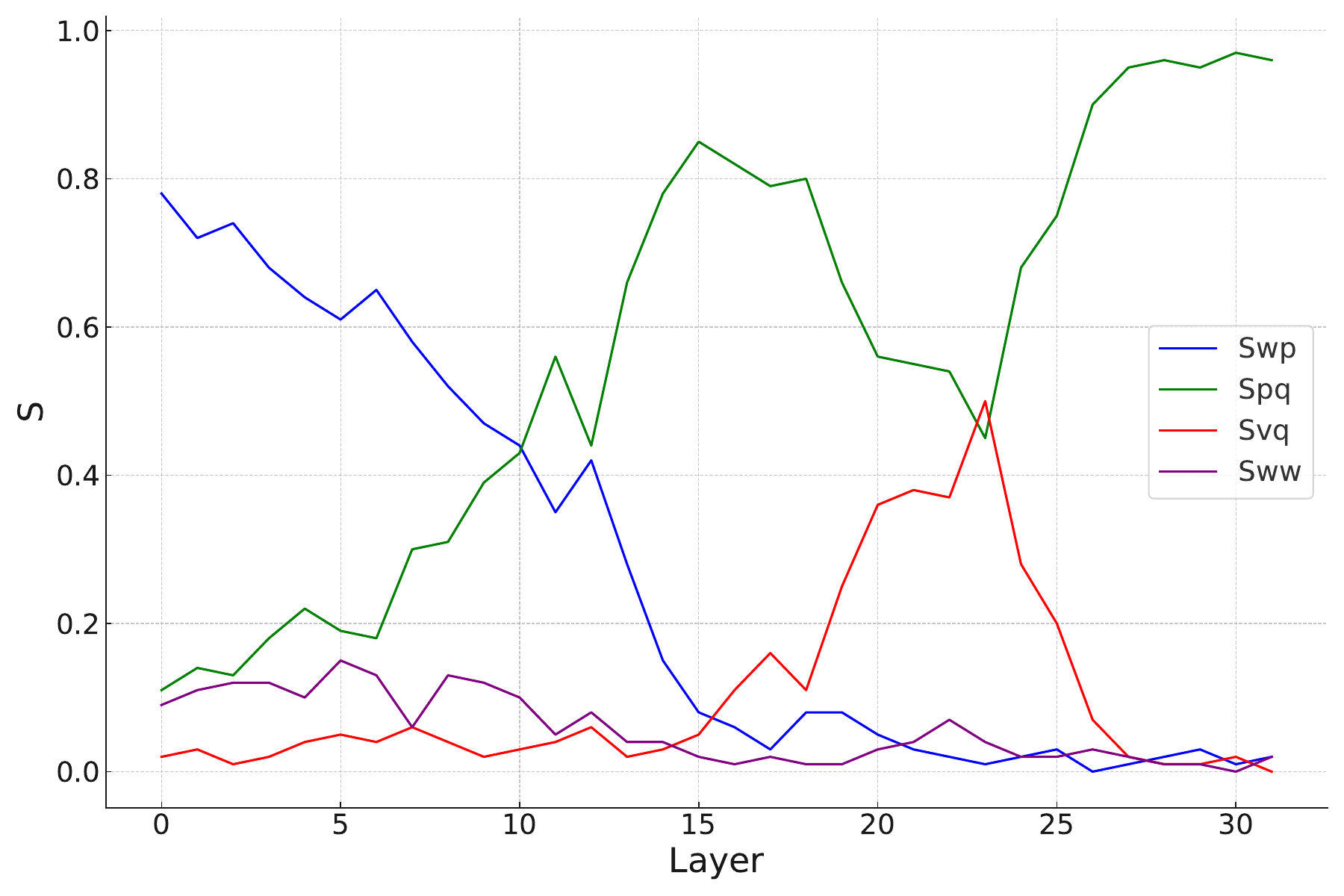}
    \vspace{-0.7cm}
    \caption{\small Information flow results on the EmoSet.}
    \label{fig:flow}
    \vspace{-0.5cm}
\end{figure}

As shown in Figure~\ref{fig:flow}, the analysis reveals varying degrees of importance in information flow within VICL across different layers and attention heads. In the shallow layers, $S_{wp}$ is highly important but diminishes as the layers progress. This suggests that image summaries are crucial in determining label words in the early stages, but their influence weakens with increasing model depth. Unlike $S_{wp}$, the importance of $S_{pq}$ increases as the layers deepen. This implies that the influence of label words on determining the target position becomes more significant in the later stages of the model. Similar to $S_{pq}$, $S_{vq}$ shows some importance in the early stages but diminishes as the layers deepen. This suggests that the influence of label words on the input image weakens as the model progresses. $S_{ww}$ remains relatively stable throughout the training process, showing no significant trend. 
Image summaries are crucial for label words aggregating information in the early stages, but the model increasingly emphasizes the relationship between label words and target position as it deepens. 

\begin{table*}[!t]\small
\centering
\setlength{\tabcolsep}{7.4pt}
\begin{tabular}{ccccccc}
\toprule
\textbf{Model}            & \textbf{Method} & \textbf{EmoSet} & \textbf{Emotion6} & \textbf{UnBiasedEmo} & \textbf{CIFAR10} & \textbf{MNIST} \\\midrule
\multirow{3}{*}{LLaVA-7B \cite{Liu2023VisualITLLAVA}}  & Zero-Shot      & 0.23            & 0.31              & 0.31                 & 0.75             & 0.85           \\
                            & ICL            & 0.32            & 0.40              & 0.38                 & 0.68             & 0.77           \\
                            & VICL           & 0.69            & 0.70              & 0.76                 & 0.84             & 0.88           \\\midrule
\multirow{3}{*}{MiniGPT-4 \cite{zhu2023minigpt}} & Zero-Shot      & 0.21            & 0.27              & 0.27                 & 0.61             & 0.84           \\
                            & ICL            & 0.28            & 0.34              & 0.36                 & 0.65             & 0.67           \\
                            & VICL           & 0.61            & 0.61              & 0.74                 & 0.76             & 0.85           \\\midrule
\multirow{3}{*}{Qwen-VL \cite{Bai2023QwenVLAF}}   & Zero-Shot      & 0.22            & 0.29              & 0.30                 & 0.74             & 0.78           \\
                            & ICL            & 0.31            & 0.39              & 0.37                 & 0.66             & 0.75           \\
                            & VICL           & 0.64            & 0.63              & 0.74                 & 0.84             & 0.85           \\\midrule
\multirow{3}{*}{LLaVA-13B \cite{Liu2023VisualITLLAVA}} & Zero-Shot      & 0.32            & 0.34              & 0.38                 & 0.79             & 0.87           \\
                            & ICL            & 0.32            & 0.52              & 0.42                 & 0.70             & 0.80           \\
                            & VICL           & \textbf{0.72}   & \textbf{0.76}     & \textbf{0.78}        & \textbf{0.87}    & \textbf{0.90}  \\\bottomrule
\end{tabular}
\vspace{-0.2cm}
\caption{\small Comparison results of LVLM on five datasets.}
\label{tab:main}
\vspace{-0.4cm}
\end{table*}

\section{Experiments}
\subsection{Experimental Settings}
\paragraph{Dataset.}
In our experiments, we used five popular datasets related to image content reasoning: EmoSet \cite{EmoSet}, Emotion6 \cite{Emotion6}, UnBiasedEmo \cite{UnBiasedEmo}, CIFAR10 \cite{CIFAR10}, and MNIST \cite{MNIST}. EmoSet, Emotion6, and UnBiasedEmo are datasets for image emotion classification, where the goal is to infer emotions based on the content of the images. EmoSet consists of 8 emotions, while Emotion6 and UnBiasedEmo consist of 6 emotions, each with multiple sub-categories. CIFAR10 and MNIST are classification datasets, where the task is to identify the object category in the images. For each of EmoSet, Emotion6, UnBiasedEmo, CIFAR10 and MNIST, we sampled 100 and 1000 samples as demonstration candidates and test sets, respectively. The metric for all test sets is accuracy.

\paragraph{Prompts.}
We have considered three distinct prompts for experimental comparison: (1) ``Zero-shot'' involves using the instruction and input image directly as the prompt without providing any demonstrations, formatted as ``\{instruction\} \{image\}''. (2) ``ICL'' (In-Context Learning) includes first retrieving demonstrations using Visual Demonstration Retrieval, then integrating these demonstrations into the prompt, formatted as ``\{instruction\} \{demonstrations\} \{image\}''. (3) ``VICL'' (Visual In-Context Learning) involves first retrieving demonstrations using Visual Demonstration Retrieval, then converting the images from the demonstrations into text using Intent-Oriented Image Summarization, concatenating this text back into the demonstrations, and finally integrating them into the prompt, formatted as ``\{instruction\} \{text demonstrations\} \{image\}''. Detailed prompt specifications are provided in Appendix~\ref{sec:prompt}.

\paragraph{Large Vision-Language Models.}
In this paper, we leverage four state-of-the-art LVLMs with various prompts to perform visual reasoning tasks. These models are LLaVA-7B \cite{Liu2023VisualITLLAVA}, MiniGPT-4 \cite{zhu2023minigpt}, Qwen-VL \cite{Bai2023QwenVLAF}, and LLaVA-13B \cite{Liu2023VisualITLLAVA}. LLaVA-7B and LLaVA-13B are based on the visual instruction tuning (VIT) technique, which aligns a frozen visual encoder and a large language model (LLM) using one projection layer. MiniGPT-4 is an open-source chatbot that fine-tunes LLaMA/Vicuna on GPT-generated multi-modal instruction-following data. Qwen-VL is a versatile vision-language model that can perform understanding image and text. We compare and analyze the in-context learning performance and capabilities of these models.

\subsection{In-Context Learning} 
We analyze the performance of LVLMs using Zero-Shot, ICL, and VICL approaches on five datasets.
The results in Table~\ref{tab:main} show the effectiveness of VICL across different LVLMs and datasets. 
VICL consistently outperforms both ICL and Zero-Shot across all models and datasets. This improvement underscores the effectiveness of VICL in enhancing the in-context learning capability of LVLMs by providing intent-oriented demonstrations. The method significantly bridges the cross-modal gap, allowing LVLMs to better understand and incorporate visual information within LLM reasoning processes.
The performance increase is more pronounced in models like LLaVA-13B, where VICL boosts performance notably compared to the baseline Zero-Shot and ICL methods. This suggests that models with higher capacity or more parameters benefit more from the VICL approach due to their stronger ability to reason with multi-modal information.

\subsection{Analysis}
\paragraph{Visual Demonstration Retrieval.}
\begin{figure}[t]
    \centering
    \includegraphics[width=1\linewidth]{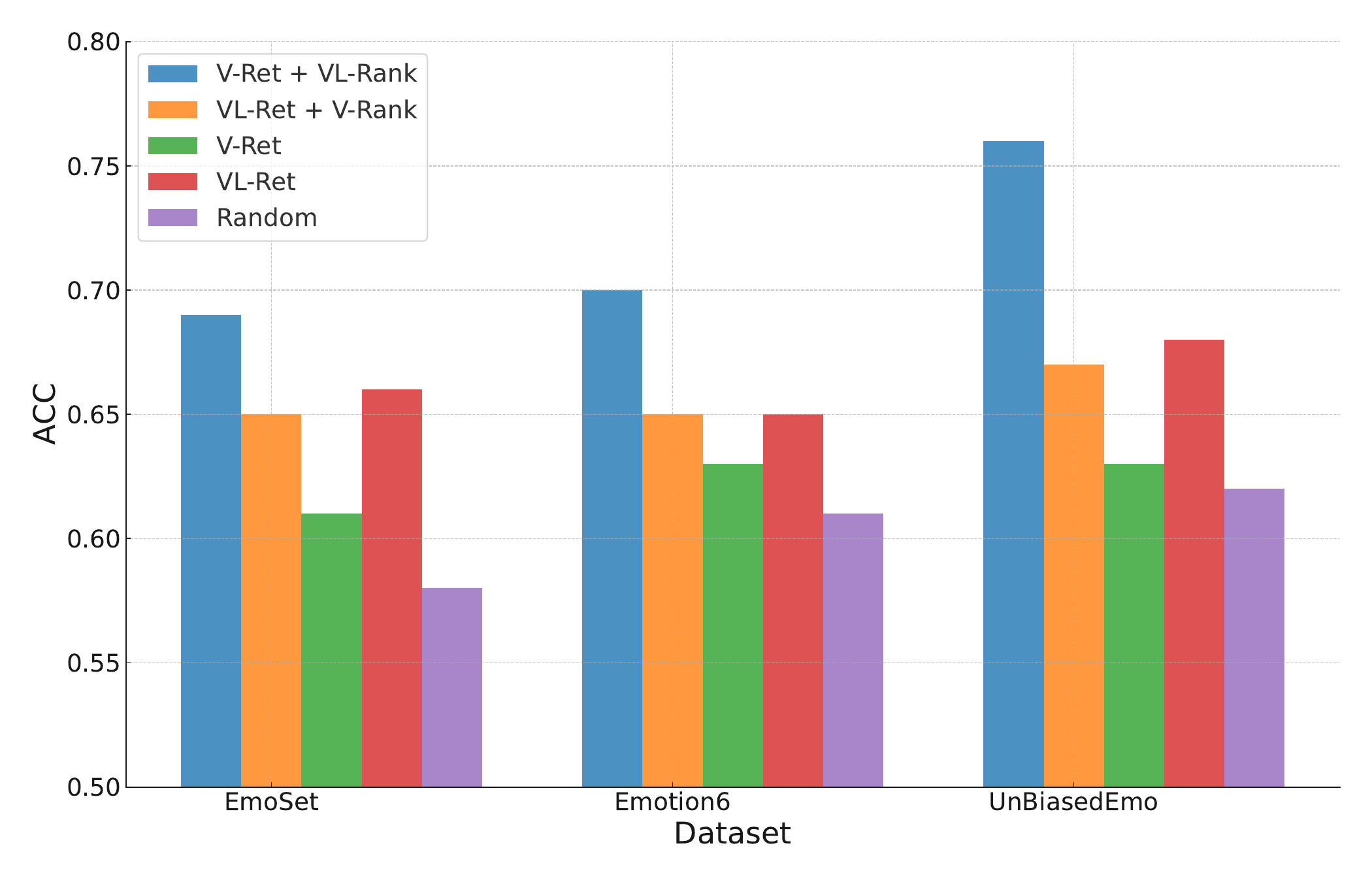}
    \vspace{-0.9cm}
    \caption{\small Diferent retrieval method comparison. ``\textit{V-Ret + VL-Rank}'' denotes the combination of ViT for retrieval and CLIP for reranking. ``\textit{VL-Ret + V-Rank}'' refers to CLIP for retrieval and ViT for reranking. ``\textit{V-Ret}'' and ``\textit{VL-Ret}'' are ViT and CLIP  alone for retrieval, respectively. ``\textit{Random}'' is random sampling.}
    \label{fig:retrieval}
    \vspace{-0.4cm}
\end{figure}
As shown in Figure~\ref{fig:retrieval}, our experiments compare retrieval and reranking strategies on three datasets. ``\textit{V-Ret + VL-Rank}'', outperforms others on the Emotion6 dataset, highlighting the benefits of broad retrieval by ViT complemented by CLIP's nuanced understanding. ``\textit{VL-Ret + V-Rank}'', maintains consistent performance across datasets but falls short of ``\textit{V-Ret + VL-Rank}'' on Emotion6 and UnBiasedEmo, suggesting ViT's unique approach may not always enhance performance. ``\textit{V-Ret}'', shows consistency but lacks the leading performance of combined approaches. ``\textit{VL-Ret}'', performs better than random sampling but lags behind two-step methods. ``\textit{Random}'' emphasizes the need for strategic retrieval and reranking.

\paragraph{Impact of Image Demonstration Number.}
\begin{figure}[t]
    \centering
    \includegraphics[width=1\linewidth]{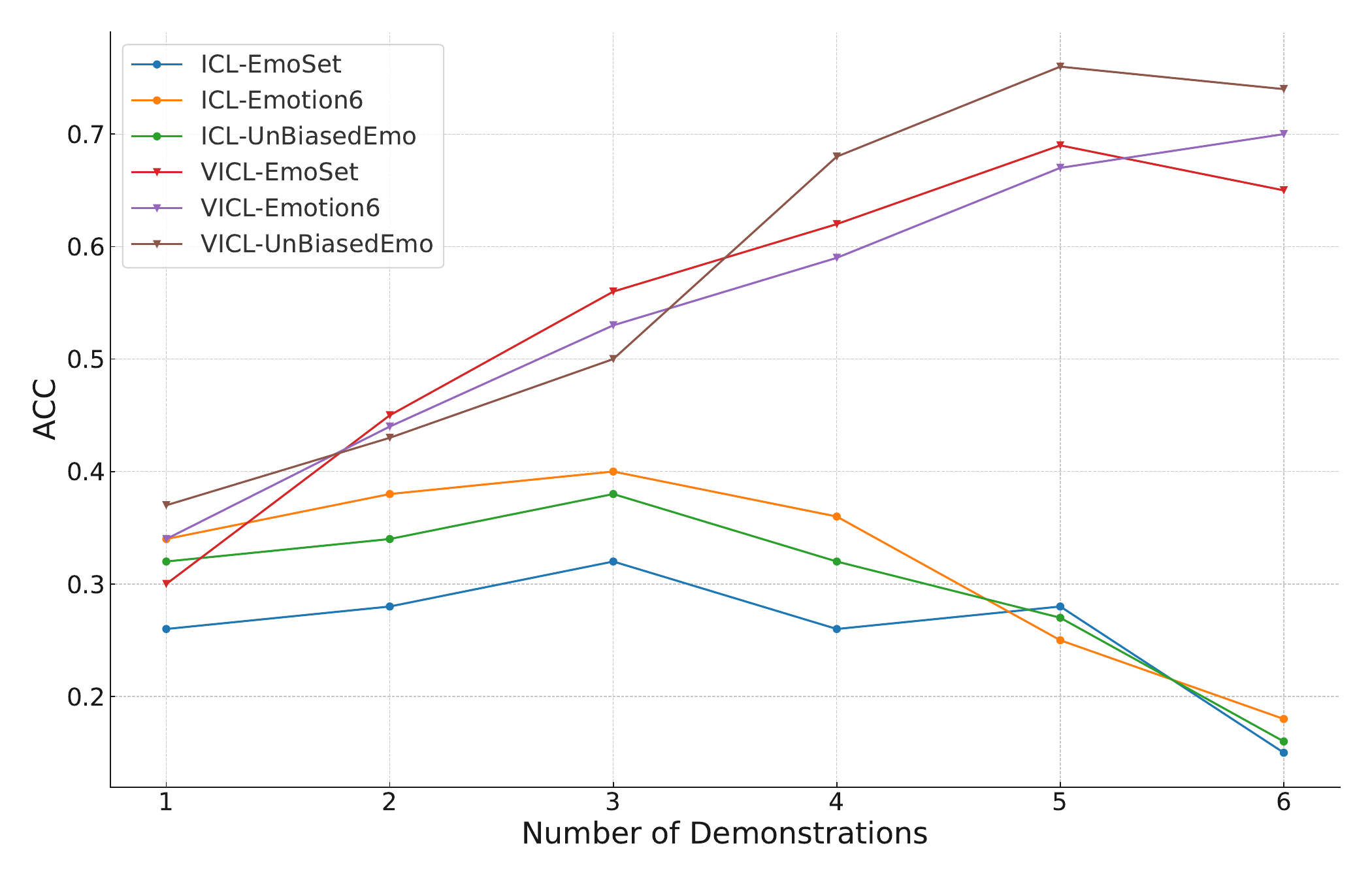}
    \vspace{-0.9cm}
    \caption{\small Impact of Image Demonstration Number.}
    \label{fig:de_num}
    \vspace{-0.4cm}
\end{figure}
Our experimental results reveal a clear impact of the number of image demonstrations on the performance of both ICL and VICL. As shown in Figure~\ref{fig:de_num}, the ICL method shows a modest increase in performance with an increasing number of demonstrations, particularly evident in the progression from one to three demonstrations. However, the performance tends to plateau or even slightly decrease beyond three demonstrations, suggesting a diminishing return on additional demonstrations. In stark contrast, the VICL method exhibits a more pronounced improvement with the increase in the number of demonstrations. This indicates that the VICL method effectively leverages additional demonstrations, translating into substantial performance gains.

\paragraph{Impact of Context Length.}
\begin{figure}[t]
    \centering
    \includegraphics[width=1\linewidth]{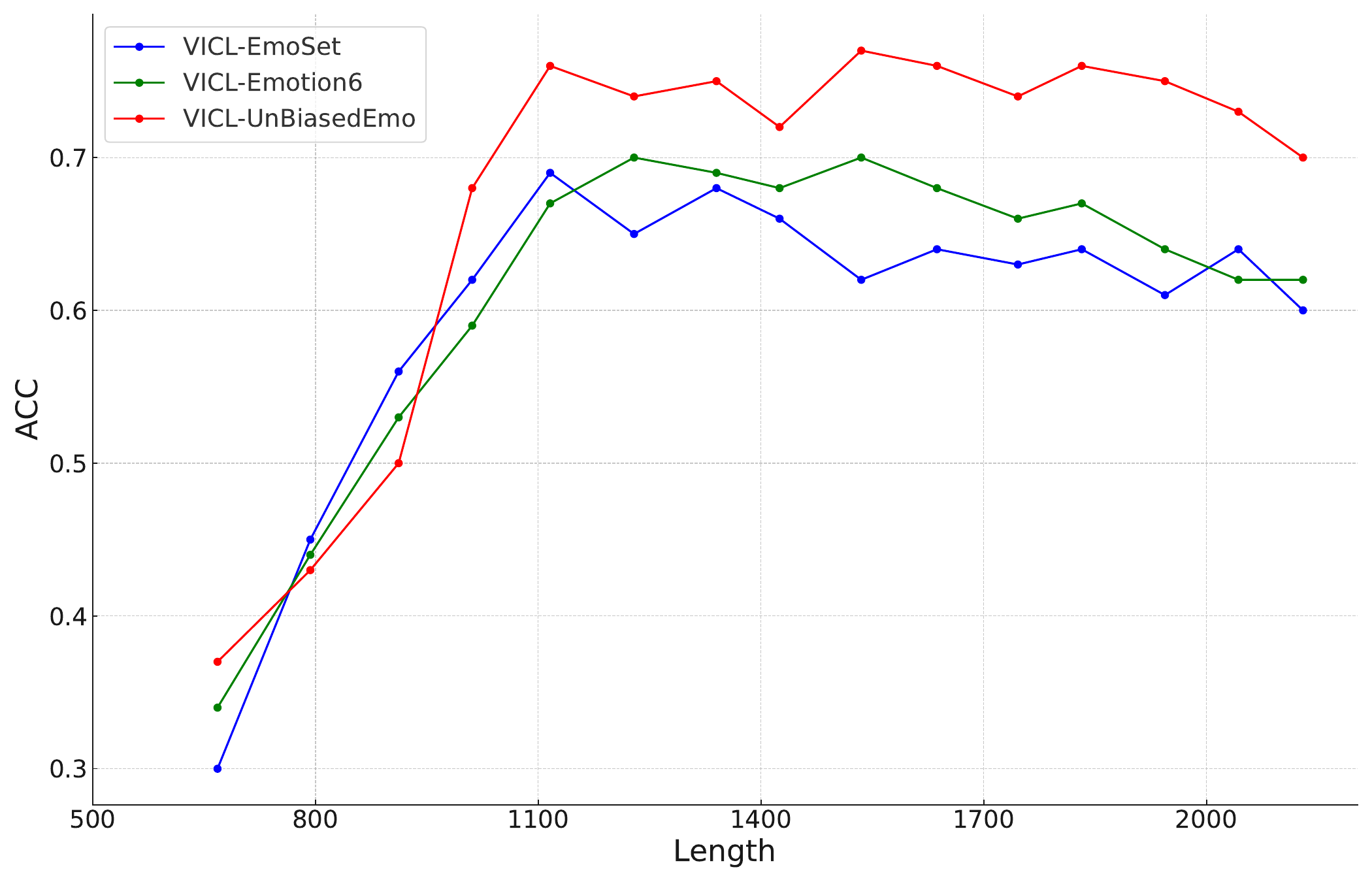}
    \vspace{-0.9cm}
    \caption{\small Impact of Context Length.}
    \label{fig:context}
    \vspace{-0.4cm}
\end{figure}
The performance dynamics of the VICL method, as observed in Figure~\ref{fig:context}, exhibit a discernible correlation with the length of the context provided. The results demonstrate an initial increase in accuracy with the expansion of context length of VICL. Particularly, on the EmoSet dataset, the accuracy ascends from 0.3 to a peak of 0.69, followed by a tapering off and slight fluctuations thereafter. A similar trend is observable in the Emotion6 and UnBiasedEmo dataset. However, beyond certain context lengths, there is a general trend of diminishing gains, or even a slight decline in accuracy. This highlights the balance between providing sufficient context for the model to leverage and avoiding an excessive amount.

\paragraph{Order of Demonstrations.}
\begin{figure}[t]
    \centering
    \includegraphics[width=1\linewidth]{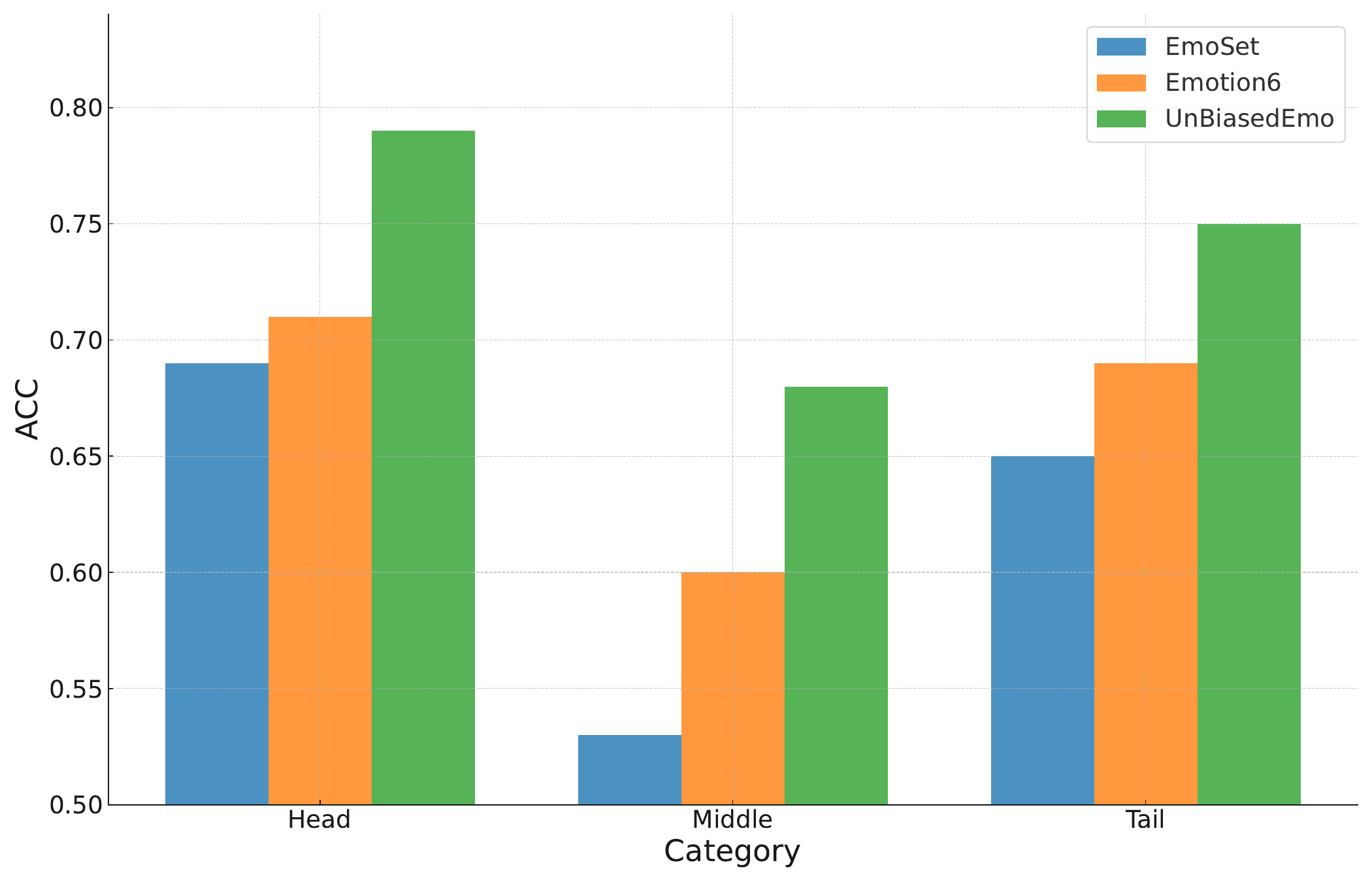}
    \vspace{-0.9cm}
    \caption{\small Impact of Demonstration Order.}
    \label{fig:order}
    \vspace{-0.4cm}
\end{figure}
We delve into the influence of the position of examples with positive labels -- labels as same as the true category of the prediction sample -- within the demonstration sequence. Following \cite{LostMiddle,ThoT}, we split the positions into three distinct sections: head, middle, and tail. As depicted in Figure~\ref{fig:order}, the head position yields the highest accuracy across all datasets, and the tail position demonstrates the next best performance. The middle position shows the least favorable performance. This trend could suggest that the model's predictions are more influenced by examples positioned at the beginning and end of the sequence. These observations underscore the significance of demonstration order in visual in-context learning.

\paragraph{Impact of Visual Summarization Method.}
\begin{table}[t]\small
\centering
\setlength{\tabcolsep}{5pt}
\begin{tabular}{cccc}
\toprule
\textbf{Method} & \textbf{EmoSet} & \textbf{Emotion6} & \textbf{UnBiasedEmo} \\\midrule
Standard        & 0.61            & 0.62              & 0.65                 \\
Task Intent     & 0.64            & 0.65              & 0.71                 \\
Image Parsing   & 0.66            & 0.68              & 0.69                 \\
IOIS            & \textbf{0.69}   & \textbf{0.70}     & \textbf{0.76}        \\\bottomrule
\end{tabular}
\vspace{-0.2cm}
\caption{\small Impact of Visual Summarization.}
\label{tab:vsum}
\vspace{-0.3cm}
\end{table}
To evaluate the effect of various visual summarization for VICL, we consider four strategies, i.e, Standard captioning, Task Intent Summarization, Image Parsing Summarization, and Intent-Oriented Image Summarization (IOIS). Details can be found in Appendix~\ref{sec:vsump}. As Table~\ref{tab:vsum} shown, task intent summarization yields a moderate increase in accuracy, demonstrating the benefit of aligning the image summary with the task. Image Parsing, including a detailed visual reasoning process, can significantly enhance performance. The IOIS method, leveraging the strengths of the previous two approaches, achieves the best performance with a notable margin. The improvement demonstrates the efficacy of integrating task intent with image parsing, suggesting that an understanding of both task and visual content is paramount.

\subsection{In-Context Unlearning}
\begin{table}[t]\small
\centering
\setlength{\tabcolsep}{1.5pt}
\begin{tabular}{ccccc}
\toprule
\multirow{2}{*}{\textbf{Method}} & \multicolumn{2}{c}{\textbf{Emotion6}}   & \multicolumn{2}{c}{\textbf{UnBiasedEmo}} \\\cmidrule(lr){2-3} \cmidrule(lr){4-5}
                                  & \textbf{Unlearning Set} & \textbf{All Set}       & \textbf{Unlearning Set}  & \textbf{All Set}       \\\midrule
Zero-Shot                         & 0.1            & 0.26          & 0.08            & 0.24          \\
ICL                               & 0.57           & 0.35          & 0.49            & 0.36          \\
VICL                              & \textbf{0.77}  & \textbf{0.69} & \textbf{0.82}   & \textbf{0.74} \\\bottomrule
\end{tabular}
\vspace{-0.2cm}
\caption{\small In-Context Unlearning.}
\label{tab:icul}
\vspace{-0.5cm}
\end{table}
We evaluate the capability of models to unlearn specific information, as shown in Table~\ref{tab:icul}. We randomly selected sub-classes from the dataset and replaced the class to build the Unlearning Set, while the entire dataset constitutes the All Set. The ``Unlearning Set'' comprises samples from five randomly selected sub-classes with labels reassigned to alternate categories and incorporated into the demonstration set and test set. Specifically, we randomly select an example corresponding to the input image's class and include it in the demonstration set. Other examples in the demonstration set are drawn from samples belonging to standard categories. This setup is designed to assess the model's ability to discard previously learned sub-class information when exposed to intentionally mislabeled examples. The performance on this set directly reflects the unlearning accuracy. Meanwhile, the ``All Set'' includes the Unlearning Set combined with additional samples from standard categories. The Zero-Shot shows the lowest performance, indicating a limited ability to disregard incorrect sub-class information based on the model's pre-existing knowledge. ICL exhibits a marked improvement in the Unlearning Set, demonstrating its ability to adapt to the new context provided by the altered demonstrations. VICL method significantly outperforms the other approaches, achieving the highest unlearning accuracy. VICL also maintains superior performance in the All Set, indicating robustness in distinguishing between correctly and incorrectly labeled samples and adjusting its inferences accordingly. 

\section{Conclusion}
This paper has introduced the integration of In-Context Learning (ICL) into Large Visual Language Models (LVLMs), addressing challenges in cross-modal interactions and the distinct representation spaces. Through Visual In-Context Learning (VICL), incorporating Visual Demonstration Retrieval, Intent-Oriented Image Summarization, and Demonstration Composition, LVLMs show enhanced performance in understanding visual and textual information. In VICL, we have not only streamlined the in-context learning process but also introduced the concept of in-context unlearning, allowing LVLMs to adjust their knowledge base dynamically without the need for retraining. Our method shows effective in improving LVLMs in processing multi-modal tasks. The extensive evaluations verify the effectiveness of our VICL method, highlighting its potential to bridge the gap between visual and linguistic modalities. 

\section*{Limitations}
This study introduces the VICL method to advance LVLMs, yet acknowledges several limitations warranting further investigation: (1) The efficacy of VICL heavily depends on the performance of LVLMs, which in turn is highly reliant on the original parameter size and the scale of training data. Further validation with larger LVLMs requires more computational resources. (2) While VICL demonstrates promise in visual reasoning tasks, its broader applications can be explored. Some Difficult tasks may require improved strategies for VICL method.

\bibliography{custom}

\appendix
\newpage
\section{Prompts}\label{sec:prompt}
There are prompts for the three methods, i.e., Zero-Shot, ICL, and VICL. 
\begin{itemize}
\item For EmoSet, Emotion6 and UnBiasedEmo dataset, the prompt for Zero-Shot is ``Question: Do you feel which emotion when seeing this image? There is an emotion category list: [\{Label List\}]. Image: \{image\}. Answer: ''. 

\item For CIFAR10 and MNIST dataset, the prompt for Zero-Shot is ``Question: What you see in this image? There is a category list: [\{Label List\}]. Image: \{image\}. Answer: ''. 

\item For EmoSet, Emotion6 and UnBiasedEmo dataset, the prompt for ICL is ``Question: Do you feel which emotion when seeing this image? There is an emotion category list: [\{Label List\}]. Image 1: \{image-1\}. Answer: \{label-1\}. Image 2: \{image-2\}. Answer: \{label-2\} $\dots$ Image N: \{image-N\}. Answer: ''.

\item For CIFAR10 and MNIST dataset, the prompt for ICL is ``Question: What you see in this image? There is a category list: [\{Label List\}]. Image 1: \{image-1\}. Answer: \{label-1\}. Image 2: \{image-2\}. Answer: \{label-2\} $\dots$ Image N: \{image-N\}. Answer: ''.

\item For EmoSet, Emotion6 and UnBiasedEmo dataset, the prompt for VICL is ``Question: Do you feel which emotion when seeing this image? There is an emotion category list: [\{Label List\}]. Image 1: \{summary-1\}. Answer: \{label-1\}. Image 2: \{summary-2\}. Answer: \{label-2\} $\dots$ Image N: \{image-N\}. Answer: ''.

\item For CIFAR10 and MNIST dataset, the prompt for VICL is ``Question: What you see in this image? There is a category list: [\{Label List\}]. Image 1: \{summary-1\}. Answer: \{label-1\}. Image 2: \{summary-2\}. Answer: \{label-2\} $\dots$ Image N: \{image-N\}. Answer: ''.
\end{itemize}

\section{Visual Summarization Prompt}\label{sec:vsump}
We investigate different visual summarization method for VICL, and the prompt for summarization as follows:
\begin{itemize}
\item The Standard captioning approach employs conventional captioning techniques, and its prompt is ``Generate a detailed description of the content depicted in the provided image.''.

\item Task Intent method enriches image descriptions with task-specific intent, and its prompt is ``Given an image and a corresponding label, generate a descriptive caption that not only describes the image content but also conveys the intention or purpose behind the depicted scene.''.

\item Image Parsing goes further by incorporating descriptions of both image observations and the reasoning process, and its prompt is ``You are presented with an image along with accompanying labels. Your task is to provide a detailed description of the image content while also explaining the observations and reasoning process behind your description.''.

\item IOIS combines features of both Task Intent and Image Parsing to provide comprehensive summaries, and its prompt is ``Generate a descriptive caption for the provided image and labels, elucidating both the visual content and the underlying purpose or intention depicted. Craft a clear and concise description that seamlessly integrates details from the image and labels, highlighting connections between visual cues and semantic meaning. Your caption should not only describe what is visible in the image but also convey the task-oriented aspect.''.

\end{itemize}

\end{document}